\documentclass[11pt,a4paper]{article}
\usepackage[hyperref]{acl2021}
\usepackage{times}
\usepackage{latexsym}

\usepackage{microtype}

\aclfinalcopy 
\def\aclpaperid{1140} 

\usepackage{graphicx}
\usepackage{tabularx}
\usepackage{adjustbox}
\usepackage{booktabs}
\usepackage{multirow}
\usepackage{amsmath}
\usepackage{cuted}
\usepackage{comment}
\usepackage{lipsum}

\title{Lightweight Cross-Lingual Sentence Representation Learning}

\author{
Zhuoyuan Mao$^{\spadesuit}$ \hspace{1em}
Prakhar Gupta$^\clubsuit$ \hspace{1em} 
Pei Wang$^\clubsuit$ \hspace{1em} \\
{\bf Chenhui Chu$^\spadesuit$} \hspace{1em}
{\bf Martin Jaggi$^\clubsuit$} \hspace{1em}
{\bf Sadao Kurohashi$^\spadesuit$}\\
$^\spadesuit$Kyoto University, Japan \hspace{1em}
$^\clubsuit$EPFL, Switzerland \\
\texttt{\{zhuoyuanmao, chu, kuro\}@nlp.ist.i.kyoto-u.ac.jp} \\
\texttt{\{prakhar.gupta, pei.wang, martin.jaggi\}@epfl.ch}
}

\date{May 20th, 2021}

\begin{document}
\maketitle
\begin{abstract}

Large-scale models for learning fixed-dimensional cross-lingual sentence representations like LASER~\cite{artetxe-schwenk-2019-massively} lead to significant improvement in performance on downstream tasks. However, further increases and modifications based on such large-scale models are usually impractical due to memory limitations. In this work, we introduce a lightweight dual-transformer architecture with just 2 layers for generating memory-efficient cross-lingual sentence representations. We explore different training tasks and observe that current cross-lingual training tasks leave a lot to be desired for this shallow architecture. To ameliorate this, we propose a novel cross-lingual language model, which combines the existing single-word masked language model with the newly proposed cross-lingual token-level reconstruction task. We further augment the training task by the introduction of two computationally-lite sentence-level contrastive learning tasks to enhance the alignment of cross-lingual sentence representation space, which compensates for the learning bottleneck of the lightweight transformer for generative tasks. Our comparisons with competing models on cross-lingual sentence retrieval and multilingual document classification confirm the effectiveness of the newly proposed training tasks for a shallow model. \footnote{\url{https://github.com/Mao-KU/lightweight-crosslingual-sent2vec}}

\end{abstract}

\section{Introduction}

Cross-lingual sentence representation models~\cite{schwenk-douze-2017-learning,DBLP:journals/jstsp/Espana-BonetVBG17,yu-etal-2018-multilingual,devlin-etal-2019-bert,chidambaram-etal-2019-learning,artetxe-schwenk-2019-massively,kim-etal-2019-learning,DBLP:journals/corr/abs-1912-12481,DBLP:conf/nips/ConneauL19,DBLP:journals/corr/abs-2007-01852,DBLP:journals/corr/abs-2008-08567} learn language-agnostic representations facilitating tasks like cross-lingual sentence retrieval (XSR) and cross-lingual knowledge transfer on downstream tasks without the need for training a new monolingual representation model from scratch. Thus, such models benefit from an increased amount of data during training and lead to improved performances for low-resource languages. 

The above-mentioned models can be categorized into two classes. On one hand, \textit{global fine-tuning} methods like mBERT~\cite{devlin-etal-2019-bert} and XLM~\cite{DBLP:conf/nips/ConneauL19} require being fine-tuned globally which results in a significant overhead of its own. On the other hand, \textit{fixed-dimensional} methods like LASER~\cite{artetxe-schwenk-2019-massively} fix the sentence representations during the pre-training phase, and subsequently the fine-tuning for specific downstream tasks without back-propagating to the pre-trained model will be extremely computationally-lite. Lightweight models have been sufficiently explored for the former group by either shrinking the model~\cite{DBLP:conf/iclr/LanCGGSS20} or training a student model~\cite{DBLP:journals/corr/abs-1910-01108,jiao-etal-2020-tinybert,reimers-gurevych-2020-making,sun-etal-2020-mobilebert}. However, the lightweight models for the latter group have not been explored before, which may have a more promising future for deploying task-specific fine-tuning onto edge devices.

\begin{table*}[t]
\begin{center}
\resizebox{\textwidth}{!}{  
\begin{tabular}{llrrrrrr}
      \toprule	
      Method&Architecture&$\mathit{d}_{h}$&$\mathit{d}_{fc}$&$\mathit{attn}_{h}$& $\mathit{Enc.}$ & $\mathit{Dec.}$ & $\mathit{Params.}$ \\
      \midrule
      mBERT~\cite{devlin-etal-2019-bert}&Transformer&768&3,072&12&12&N/A&110M\\
      LASER~\cite{artetxe-schwenk-2019-massively}&Bi-LSTM&512$\times$2&N/A&N/A&5&5&154M\\
      T-LASER~\cite{DBLP:journals/corr/abs-2008-08567}&Transformer&1,024&4,096&16&6&1&246M\\
      \midrule
      Ours&Transformer&\textbf{512}&\textbf{1,024}&\textbf{8}&\textbf{2}&N/A&\textbf{30M}\\
      \bottomrule
\end{tabular}
}

\caption{\textbf{Model sizes of related work and ours.} Our work mainly focuses on the comparison with previous fixed-dimensional methods like LASER, T-LASER, etc. $\mathit{d_h}$, $\mathit{d_{fc}}$, $\mathit{attn_h}$, $\mathit{Enc.}$, $\mathit{Dec.}$, $\mathit{Params.}$ denote dimension of the hidden state, dimension of the feed-forward hidden state, number of the attention heads, number of the encoder layers, number of the decoder layers, and number of the parameters respectively.}
\label{parameter2}

 \end{center}
\end{table*}

In this work, we propose a variety of training tasks for a lightweight cross-lingual sentence model while retaining the robustness. To improve the computational efficiency, we utilize a lightweight dual-transformer architecture with just 2 layers,  significantly decreasing the memory consumption and accelerating the training to further improve the efficiency. Our model uses significantly less number of parameters compared to both global fine-tuning methods like mBERT, and fixed-dimensional representation methods like LASER, and T-LASER~\cite{DBLP:journals/corr/abs-2008-08567} (see Table \ref{parameter2}). 

Given a fixed training-set and model architecture, the robustness of the sentence representation is dependent on the training task. It is much more difficult for a lightweight model to learn robust representations merely with existing generative tasks (see Section~\ref{relatedwork} and Section~\ref{LMS}), which could be attributed to its smaller size. In order to ameliorate this problem, we redesign a cross-lingual language model by combining the single-word masked language model (SMLM) with cross-lingual token-level reconstruction (XTR). Furthermore, we introduce two contrastive learning methods as auxiliary tasks to compensate for the learning bottleneck of lightweight transformer for generative tasks. Following the state-of-the-art fixed-dimensional model LASER, we proceed to learn cross-lingual sentence representations from parallel sentences, where we employ 2-layer dual-transformer encoders to shrink the model architecture. By introducing the above-stated training tasks, we establish a computationally-lite framework for training cross-lingual sentence models.\footnote{\citet{PeiWang} attempted to train cross-lingual sentence representation with the lightweight dual-transformer architecture. We continue their exploration by introducing new training tasks that result in better performance, and our codes are based on theirs.}

We evaluate the learned sentence representations on cross-lingual tasks including multilingual document classification (MLDoc)~\cite{schwenk-li-2018-corpus} and XSR. Our results confirm the ability of our lightweight model to yield robust sentence representations. We also do a systematic study on the performance of our model in an ablative manner. The contributions of this work can be summarized as follows:
\begin{itemize}
    \item We implement fixed-dimensional cross-lingual sentence representation learning in a lightweight model, achieving improved training efficiency and competitive performance of the learned sentence representations.
    \item Our proposed novel generative and contrastive tasks 
    allow cross-lingual sentence representation efficiently trainable by the lightweight model. The contribution from each task is empirically analyzed.
\end{itemize}

\section{Related Work}
\label{relatedwork}



A majority of training tasks for learning fixed-dimensional cross-lingual sentence representations can be ascribed to one of the following 2 categories: generative or contrastive. In this section, we revisit the previous work in these 2 categories, which is crucial for designing a cross-lingual representation model.

\noindent \textbf{Generative Tasks.}
Generative tasks measure a generative probability between predicted tokens and real tokens by training a language model. BERT-style MLM~\cite{devlin-etal-2019-bert} masks and predicts contextualized tokens within a given sentence. For the cross-lingual scenario, cross-lingual supervision is implemented by shared cognates and joint training \cite{devlin-etal-2019-bert}, 
concatenating source sentences in multiple languages~\cite{DBLP:conf/nips/ConneauL19,conneau-etal-2020-unsupervised} or explicitly predicting the translated token~\cite{ren-etal-2019-explicit}. The [CLS] embedding or pooled embedding of all the tokens is introduced as the classifier embedding, which can be used as sentence embedding for sentence-level tasks~\cite{reimers-gurevych-2019-sentence}. Sequence to sequence methods~\cite{schwenk-douze-2017-learning,DBLP:journals/jstsp/Espana-BonetVBG17,artetxe-schwenk-2019-massively,DBLP:journals/corr/abs-2008-08567} autoregressively reconstruct the translation of the source sentence. The intermediate state between the encoder and the decoder are extracted as sentence representations. Particularly, the cross-lingual sentence representation quality of LASER~\cite{artetxe-schwenk-2019-massively} benefits from a massively multilingual machine translation task covering 93 languages. In our work, we revisit the BERT-style training tasks and introduce a novel generative loss enhanced by KL-Divergence based token distribution prediction. Our proposed generative task performs effectively for the lightweight dual-transformer framework while other generative tasks should be implemented via a large-capacity model.

\noindent \textbf{Contrastive Tasks.}
Contrastive tasks measure (contrast) the similarities of sample pairs in the representation space. Negative sampling, which is a typical feature of the contrastive methods is first introduced in the work of word representation learning~\cite{DBLP:conf/nips/MikolovSCCD13}. Subsequently, contrastive tasks gradually emerged in many NLP tasks in various ways: negative sampling in knowledge graph embedding learning~\cite{DBLP:conf/nips/BordesUGWY13,DBLP:conf/aaai/WangZFC14}, next sentence prediction in BERT~\cite{devlin-etal-2019-bert}, token-level discrimination in ELECTRA~\cite{DBLP:conf/iclr/ClarkLLM20}, sentence-level discrimination in DeCLUTR~\cite{DBLP:journals/corr/abs-2006-03659}, and hierarchical contrastive learning in HICTL~\cite{DBLP:journals/corr/abs-2007-15960}. For the cross-lingual sentence representation training, typical ones include using correct and wrong translation pairs introduced by~\citet{guo-etal-2018-effective,DBLP:conf/ijcai/YangAYGSCSSK19,chidambaram-etal-2019-learning,DBLP:journals/corr/abs-2007-01852} or utilizing similarities between sentence pairs by introducing a regularization term~\cite{yu-etal-2018-multilingual}. As another advantage, contrastive methods have proven to be more efficient than generative methods~\cite{DBLP:conf/iclr/ClarkLLM20}. Inspired by previous work, for our lightweight model, we propose a robust sentence-level contrastive task by leveraging similarity relationships arising from translation pairs.

\section{Methodology}

\begin{figure*}[t]
\begin{center}
\includegraphics[width=\linewidth]{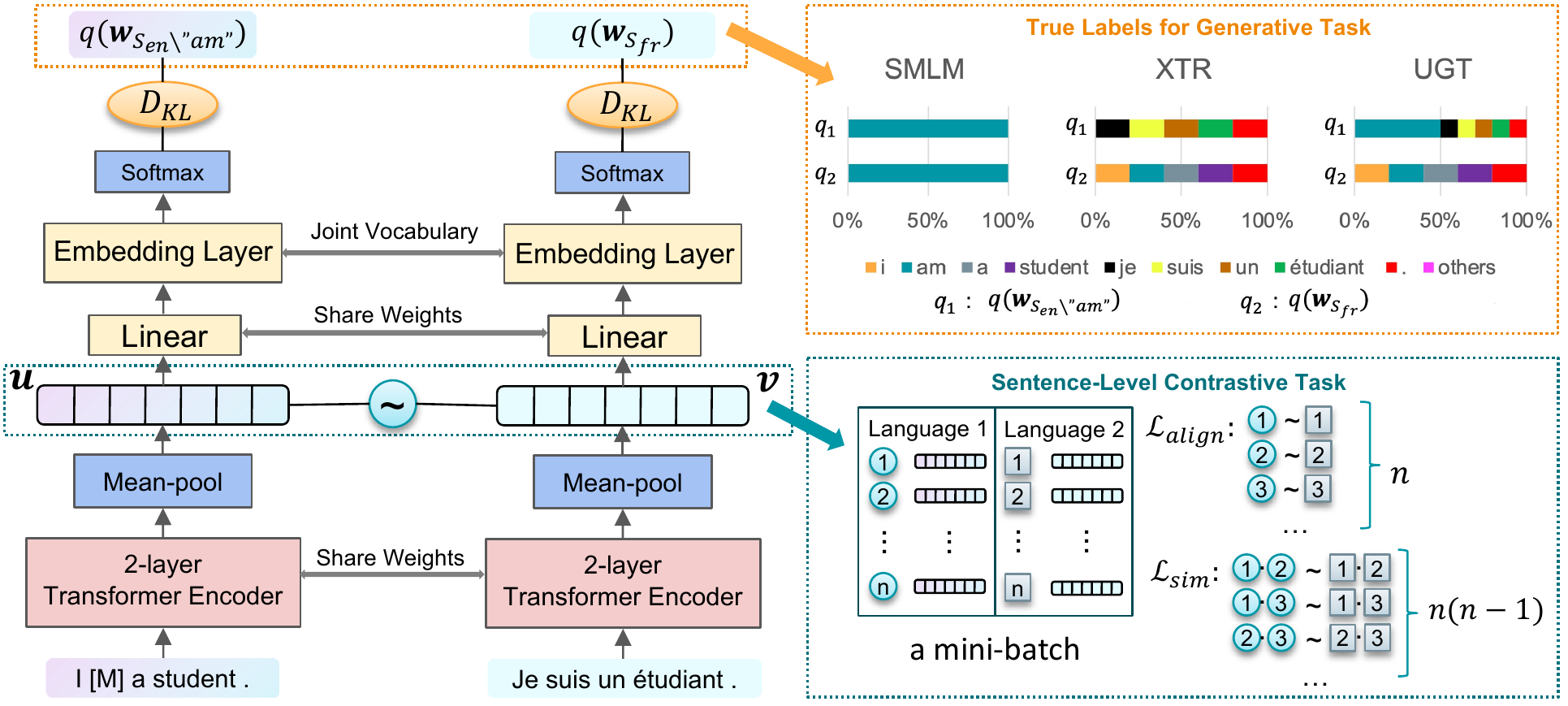}
\caption{\textbf{Architecture of the proposed model (left), proposed unified generative task (top right), and proposed sentence-level contrastive task (bottom right).} In the left sub-figure, [M] denotes the masked token introduced by SMLM. Hidden states $u$ and $v$ are 512-dimensional sentence representations for the sentence-level contrastive task and for downstream tasks. In the top right sub-figure, SMLM is inspired by~\citet{DBLP:journals/corr/abs-1912-12481}; XTR and UGT are our proposed methods. $q_1$ and $q_2$ respectively denote 2 distributions at the top of left sub-figure, the token distributions that we introduce as labels for the model to learn. In the bottom right sub-figure, n denotes the size of a mini-batch. $\bigcirc$ and $\Box$ represent language $l_1$ and $l_2$, respectively. $i$ in $\bigcirc$ indicates the sentence representation of the i-th $l_1$ sentence in the mini-batch and same for $j$ in $\Box$.}
\label{tinysent}
\end{center}
\end{figure*}

We perform cross-lingual sentence representation learning by a lightweight dual-transformer framework. Concerning the training tasks, we propose a novel cross-lingual language model, which combines SMLM and XTR. Moreover, we introduce two sentence-level self-supervised learning tasks (sentence alignment and sentence similarity losses) to leverage robust parallel level supervision to better conduct the cross-lingual sentence representation space alignment.

\subsection{Architecture}
We employ the dual transformer sharing parameters without any decoder as the basic unit to encode parallel sentences respectively, to avoid the loss in efficiency caused by the presence of a decoder. 
Unlike XLM~\cite{DBLP:conf/nips/ConneauL19}, we utilize a dual model architecture rather than a single transformer to encode sentence pairs, because it can force the encoder to capture more cross-lingual characteristics~\cite{reimers-gurevych-2019-sentence, DBLP:journals/corr/abs-2007-01852}. Moreover, we decrease the number of layers and embedding dimension to accelerate the training phase, as shown in Table~\ref{parameter2}.

The architecture of the proposed method is illustrated in Figure~\ref{tinysent} (left). We build sentence representations on the top of 2-layer transformer~\cite{DBLP:conf/nips/VaswaniSPUJGKP17} encoders by a mean-pooling operation from the final states of all the positions within a sentence. Pre-trained sentence representations for downstream tasks are denoted by $\boldsymbol{u}$ and $\boldsymbol{v}$, which are used to compute the loss for the sentence-level contrastive task. Moreover, we add a fully-connected layer before computing the loss of the cross-lingual language model inspired by~\citet{DBLP:conf/icml/ChenK0H20}. This linear layer can enhance our lightweight model by a nontrivial margin, because the hidden state for computing loss for the generative task is far different from the sentence presentation we aim to train.
Two transformer encoders and linear layers share parameters, which has been proved effective and necessary for cross-lingual representation learning~\cite{conneau-etal-2020-emerging}.

\subsection{Generative Task}
\noindent\textbf{SMLM.}
SMLM is proposed by~\citet{DBLP:journals/corr/abs-1912-12481}, which is a variant of the standard MLM in BERT \cite{devlin-etal-2019-bert}. SMLM can enforce the monolingual performance, because the prediction of a number of masked tokens in MLM is too complicated for the shallow transformer encoder to learn.\footnote{A detailed comparison between SMLM and MLM under our lightweight model setting is conducted (see Section~\ref{LMS}).} Inspired by this, we implement SMLM by a dual transformer architecture. The transformer encoder for language $l_1$ predicts a masked token in a sentence in $l_1$ as the monolingual loss. The language $l_2$ encoder sharing all the parameters with $l_1$ encoder predicts the same masked token by the corresponding sentence (translation in $l_2$) as the cross-lingual loss, as shown in Figure~\ref{tinysent} (top right). Specifically, for a parallel corpus $\boldsymbol{C}$ and language $l_1$ and $l_2$, the loss of SMLM computed from $l_1$ encoder $\boldsymbol{E_{l_1}}$ and $l_2$ encoder $\boldsymbol{E_{l_2}}$ is formulated as:

\begin{equation}
\begin{split}
\mathcal{L}_{SMLM}= 
\sum_{\substack{S \in \boldsymbol{C} \\ l, l^{\prime} \in \{l_{1}, l_{2}\} \\ \ l \neq l^{\prime}}} \big\{ -\log\left(P(w_{t} | S_{{l} \backslash\left\{w_{t}\right\}}; \boldsymbol{\theta})\right)~ \\[-1em]
-\log\left(P(w_{t} | S_{l^{\prime}}; \boldsymbol{\theta})\right) \big\}
\end{split}
\label{smlm}
\end{equation}
where $w_{t}$ is the word to be predicted, $S_{{l_1} \backslash\left\{w_{t}\right\}}$ is a sentence in which $w_{t}$ is masked, $S=(S_{l_1}, S_{l_2})$ denotes a parallel sentence pair, $\boldsymbol{\theta}$ represents the parameters to be trained in $\boldsymbol{E_{l_1}}$ and $\boldsymbol{E_{l_2}}$, and the classification probability $P$ is computed by Softmax on the top of the embedding layer.

\noindent\textbf{XTR.}
Inspired by LASER, we also use a reconstruction loss. However, introducing a decoder to implement the translation loss like LASER will increase the computational overhead associated with our model, which contradicts with our objective to design a computationally-lite model architecture.

To implement the reconstruction loss with just the encoder, we propose a XTR loss by which we jointly enforce the encoder to reconstruct the word distribution of corresponding target sentence as shown by $q$ in Figure~\ref{tinysent} (top right). 
Specifically, we utilize the following KL-Divergence based formulation as the training loss:

\begin{align}
\resizebox{\linewidth}{!}{ 
$\mathcal{L}_{XMLM} =
\sum_{\substack{S \in \boldsymbol{C} \\ l, l^{\prime} \in \{l_{1}, l_{2}\} \\ \ l \neq l^{\prime}}} \big\{ 
\mathcal{D}_{KL}\left(q\left(\mathbf{w}_{S_{l}}\right)\parallel p\left(\mathbf{h}_{S_l};\boldsymbol{\theta}\right)\right)  \nonumber$}\\
\resizebox{0.55\linewidth}{!}{ 
$+ \mathcal{D}_{KL}\left(q\left(\mathbf{w}_{S_{l^{\prime}}}\right)\parallel p\left(\mathbf{h}_{S_{l^{\prime}}};\boldsymbol{\theta}\right)\right) \big\}$}
\label{xmlm}
\end{align}
where $\mathcal{D}_{KL}$ denotes KL-Divergence based loss, $p\left(\mathbf{h}_{S_{l}};\boldsymbol{\theta}\right)$ represents the hidden state on the top of encoder $\boldsymbol{E_{l}}$ as shown in Figure~\ref{tinysent} (left) under the input $S_{l}$, and $\mathbf{w}_{S_l}$ indicates the set that contains all the tokens in the vocabulary where the probability for each token are defined using tokens within $S_{l^{\prime}}$. We utilize discrete uniform distribution for the tokens in target language to define $q$ for $\mathbf{w}_{S_l}$. Specifically,  $q\left(\mathbf{w}_{S_{l}}\right)$ is defined as:
\begin{equation}
q\left(w_i\right)=\left\{
\begin{aligned}
\frac{N_{w_i}}{\left\|S_{l^{\prime}}\right\|}&,& \hspace{1em} w_i\in S_{l^{\prime}}\\
0&,& \hspace{1em} w_i\notin S_{l^{\prime}}
\end{aligned}
\right.
\label{kl}
\end{equation}
where $N_{w_i}$ indicates the number of words $w_i$ in sentence $S_{l^{\prime}}$ and $\left\|S_{l^{\prime}}\right\|$ indicates the length of $S_{l^{\prime}}$.\footnote{We set all the $N_{w_i}$ to be 1 in the current implementation. Word frequency will be taken into consideration for the generative task in future work.}

\noindent\textbf{Unified Generative Task (UGT).}
Finally, we unify SMLM (Eq.~(\ref{smlm})) and XTR (Eq.~(\ref{xmlm})) by redefining the label distribution $q\left(\mathbf{w}_{S_{l}}\right)$ for KL-Divergence based loss. As shown in Figure~\ref{tinysent} (top right), the model is forced to learn under the supervision of a biased cross-lingual probability distribution of tokens. It is formulated the same as Eq.~(\ref{kl}) if the token $w_t$ is masked from $S_{l^{\prime}}$, else if $w_t$ is masked within $S_{l}$:

\begin{equation}
q\left(w_i\right)=\left\{
\begin{aligned}
\frac{N_{w_i}}{2\left\|S_{l^{\prime}}\right\|}&,& \hspace{1em} w_i\in S_{l^{\prime}}\\
1/2&,& \hspace{1em} w_i = w_t\\
0&,& \hspace{1em} others
\end{aligned}
\right.
\label{combined}
\end{equation}

\subsection{Sentence-Level Contrastive Task}

Meanwhile, as shown in Figure~\ref{tinysent} (bottom right), we introduce two auxiliary similarity-based training tasks to strengthen sentence-level supervision. We construct these two assisting tasks on the basis of mean pooled sentence representations, aiming to capture sentence similarity information across languages. 

Inspired by \citet{guo-etal-2018-effective,DBLP:conf/ijcai/YangAYGSCSSK19,DBLP:journals/corr/abs-2007-01852}, we propose a sentence alignment loss. The sentence alignment loss aims to force the transformer model to recognize the sentence pair, where one sentence is the translation of the other. One positive and other negative samples contribute to the gradient update in a single batch, which provides contrastive training patterns for the model training. 
For contrastively discriminating positive and negative samples, 
we use $(batch size - 1) \times 2$ negative samples.\footnote{For each language, there are $batch size - 1$ negative samples. Note that this contrastive task is different from those in~\citet{DBLP:conf/ijcai/YangAYGSCSSK19} and~\citet{DBLP:journals/corr/abs-2007-01852}, where they utilize cosine similarity while we directly use the inner product to accelerate the model.} This indicates all the sentences within a batch except the positive one will be negative samples.


More precisely, assuming the mean pooled sentence representations of $S_{l_1}$ and $S_{l_2}$ are $\boldsymbol{u}(S_{l_1})$ and $\boldsymbol{v}(S_{l_2})$. Assume that $\boldsymbol{B}_i$ is a specific batch of several paired sentences, $\boldsymbol{u}_{ij}$ and $\boldsymbol{v}_{ij}$ respectively indicate the representation of j-th sentence $S^{(j)}=(S_{l_1}^{(j)}, S_{l_2}^{(j)})$ in language $l_1$ and $l_2$ within batch $\boldsymbol{B}_i$. Note that the masked token $w_t$ is omitted in the following equations. 
The above-proposed in-batch sentence alignment loss to align sentence pairs is defined as:
\begin{equation}
\begin{split}
\mathcal{L}_{align} =
-\sum_i\sum_j(\log\frac{\exp{(\boldsymbol{u}_{ij}^\top\boldsymbol{v}_{ij})}}{\sum_{k} \exp{(\boldsymbol{u}_{ij}^\top\boldsymbol{v}_{ik})}} \\ + \log\frac{\exp{(\boldsymbol{u}_{ij}^\top\boldsymbol{v}_{ij})}}{\sum_{k} \exp{(\boldsymbol{u}_{ik}^\top\boldsymbol{v}_{ij})}} )
\label{align}
\end{split}
\end{equation}
where $S^{(k)}, S^{(j)} \in \boldsymbol{B}_i$.

We further introduce a sentence similarity loss to better align similarities for all the sentence pairs throughout a batch. By constructing these similarity-based sentence-level contrastive tasks, we hope that it can force the sentence representations to be competent for sentence-level alignment downstream tasks.
Specifically, in-batch sentence similarity loss, $\mathcal{L}_{sim}$ is formulated as:

\begin{equation}
\begin{split}
\mathcal{L}_{sim} =
- \sum_i\sum_j\log\cos\left\{
\frac{\pi}{2}(
\frac{\exp{(\boldsymbol{u}_{ij_1}^\top\boldsymbol{u}_{ij_2})}}{\sum_{k} \exp{(\boldsymbol{u}_{ij_1}^\top\boldsymbol{u}_{ik})}} \right. \\ \left.
- \frac{\exp{(\boldsymbol{v}_{ij_1}^\top\boldsymbol{v}_{ij_2})}}{\sum_{k} \exp{(\boldsymbol{v}_{ij_1}^\top\boldsymbol{v}_{ik})}})
\right\}
\end{split}
\label{sim2}
\end{equation}
where $S^{(k)},S^{(j)} \in \boldsymbol{B}_i$.\footnote{With regard to Eq.~\ref{sim2}, $\log\cos$ is employed for implementing a regression loss because we focused on the hidden states after Softmax that indicate the probabilities. We will consider using MSE loss on the states before Softmax in future exploration.}

In summary, Eq.~(\ref{align}) optimizes a loss for the contrastive task by discriminating correct translation from others for a given sentence, as shown in Figure~\ref{tinysent} ($\mathcal{L}_{align}$ in bottom right). Eq.~(\ref{sim2}) aligns the cross similarities between every sentence pairs within a batch, as shown in Figure~\ref{tinysent} ($\mathcal{L}_{sim}$ in bottom right). The similarity score matrix generated by the inner product between sentence pairs in a batch will be trained to be a symmetrical matrix with diagonal elements approximate to 1 after the Softmax operation.

\subsection{Weighted Loss for Generative and Contrastive Tasks}
We jointly minimize the loss of the generative task and two auxiliary contrastive tasks with the weight combination of $(1,2,2)$:\footnote{We assign a bigger weight for contrastive tasks according to the task discrepency between the generative task and contrastive tasks introduced by sentence pair similarities.}
\begin{equation}
\begin{split}
\mathcal{L}(\omega_0, \omega_1, \omega_2) =
\mathcal{L}_{XMLM} + 2\mathcal{L}_{align} + 2\mathcal{L}_{sim}
\end{split}
\end{equation}
where $\mathcal{L}_{XMLM}$ denotes the loss of Eq.~(\ref{xmlm}) and the label distribution for KL-Divergence based loss is the unified reconstruction distribution formulated by Eq.~(\ref{combined}). $\mathcal{L}_{align}$ and $\mathcal{L}_{sim}$ represent the losses in Eq.~(\ref{align}) and Eq.~(\ref{sim2}), respectively.



\section{Experiments}

We evaluate our cross-lingual sentence representation models by cross-lingual document classification and bitext mining for these 2 main downstream tasks belong to 2 groups: unrelated and related to the training task. For the former, we select MLDoc~\cite{schwenk-li-2018-corpus} to evaluate the classifier transfer ability of the cross-lingual model, while for the latter we conduct sentence retrieval on another parallel dataset Europarl\footnote{\url{https://www.statmt.org/europarl/}} to evaluate the performance of our models.

\subsection{Configuration Details}

\begin{table}[h!]
\begin{center}
\resizebox{\columnwidth}{!}{  
\begin{tabular}{lrrrr}
      \toprule	
      Language Pair&en-fr&en-de&en-es&en-it\\
      \midrule
      Raw&51.3M&36.9M&39.0M&22.1M\\
      Filtered&37.8M&29.6M&32.8M&17.3M\\
      \bottomrule
\end{tabular}
}

\caption{\textbf{Training data overview.} Number of raw and filtered parallel sentences from ParaCrawl v5.0.} 
\label{data}

 \end{center}
\end{table}

\begin{table*}[t!]
\begin{center}
\resizebox{\linewidth}{!}{  
\begin{tabular}{lrrrrrrrrrr}
      \toprule	
     \multirow{2}{*}{Method} &\multicolumn{2}{c}{en-fr}&\multicolumn{2}{c}{en-de}&\multicolumn{2}{c}{en-es}&\multicolumn{2}{c}{en-it}&\multirow{2}{*}{Avg.}\\
      &$\rightarrow$&$\leftarrow$&$\rightarrow$&$\leftarrow$&$\rightarrow$&$\leftarrow$&$\rightarrow$&$\leftarrow$&\\
      \toprule
      
      \textit{fixed-dimensional word representation methods}\\
      MultiCCA + CNN~\cite{schwenk-li-2018-corpus}&72.4&64.8&81.2&56.0&72.5&74.0&69.4&53.7&68.0\\
      Bi-Sent2Vec~\cite{DBLP:journals/corr/abs-1912-12481}   &81.6&82.2&86.5&79.2&74.0&71.5&\textbf{75.0}&72.6&77.8\\
      \hline
      \textit{fixed-dimensional sentence representation methods}\\
      \citet{yu-etal-2018-multilingual}&80.8&81.0&80.2&77.1&74.1&74.1&70.8&74.8&76.6\\
      LASER~\cite{artetxe-schwenk-2019-massively}&78.0	&80.1&	86.3&	\textbf{80.8}&	79.3	&69.6	&70.2&	74.2&77.3\\
      T-LASER~\cite{DBLP:journals/corr/abs-2008-08567} &70.7&78.2&86.8&79.0&71.4&74.5&68.7&76.0& 75.7\\
      \midrule
      \textbf{Ours} &\textbf{85.1}&\textbf{82.4}&\textbf{88.8}&\textbf{80.8}&\textbf{80.8}&\textbf{79.2}&74.3&\textbf{79.9}&\textbf{81.4}\\
      \hline
      \hline
      \textit{reference: global fine-tuning style methods}\\
      mBERT~\cite{devlin-etal-2019-bert}&83.0	&-	&82.4&	-	&75.0&	-	&68.3&	-&-\\
      MultiFit~\cite{eisenschlos-etal-2019-multifit} &89.4&-&91.6&-&79.1&-&76.0&- &-\\
      \bottomrule
\end{tabular}
}

\caption{\textbf{MLDoc benchmark results (zero-shot scenario).} We compare our models primarily with fixed-dimensional models in which Bi-Sent2vec and LASER are state-of-the-art bag-of-words based and contextual sentence representation models, respectively. We also compare with global fine-tuning style methods here for reference. Each result is the mean value of 5 runs.}
\label{mldoc}

 \end{center}
\end{table*}

\begin{table*}[t!]
\begin{center}
\resizebox{\linewidth}{!}{  
\begin{tabular}{lr rrrr rrr r}
      \toprule
      \multirow{2}{*}{Method}&\multicolumn{2}{c}{en-fr}&\multicolumn{2}{c}{en-de}&\multicolumn{2}{c}{en-es}&\multicolumn{2}{c}{en-it}&\multirow{2}{*}{Avg.}\\
      &$\rightarrow$&$\leftarrow$&$\rightarrow$&$\leftarrow$&$\rightarrow$&$\leftarrow$&$\rightarrow$&$\leftarrow$&\\
      \toprule
      \textit{bilingual representation methods}\\
      Bi-Vec~\cite{luong-etal-2015-bilingual} &81.6&	83.4&	71.6&	68.1&	81.6&	83.4&	74.2	&72.4&77.0\\
      Bi-Sent2Vec ~\cite{DBLP:journals/corr/abs-1912-12481} &87.4&	87.8&	84.0&	84.2&	89.6	&89.7&	\textbf{87.6}&	\textbf{87.9}&87.3\\
      \midrule
      \textbf{Ours} & \textbf{90.2}&\textbf{90.8}	&\textbf{86.3}	&\textbf{86.9}&\textbf{90.7}&\textbf{91.2}&86.9&87.6&\textbf{88.8}\\
      \hline
      \hline
      \textit{multilingual representation methods} \\
      TransGram~\cite{coulmance-etal-2015-trans}&80.4&81.6&72.7&69.1&83.8&82.7&77.9&77.2&78.2\\
      LASER~\cite{artetxe-schwenk-2019-massively} & 95.3	& 94.7	& 94.6	&94.3&	94.5	&94.1&	95.6	& 95.6 &94.8\\
      \bottomrule
\end{tabular}
}
\caption{\textbf{Cross-lingual sentence retrieval results.} We report P@1 scores of 2,000 source queries when searching among 200k sentences in the target language. Here \textit{global fine-tuning style methods} are not considered, because they require training data to be fine-tuned. Best performances among bilingual representation methods are in bold.}
\label{csr}

 \end{center}
\end{table*}



We build our PyTorch implementation on top of HuggingFace's Transformers library ~\cite{wolf-etal-2020-transformers}. Training data is composed of the ParaCrawl\footnote{\url{http://opus.nlpl.eu/ParaCrawl-v5.php}}~\cite{banon-etal-2020-paracrawl} v5.0 datasets for each language pair. We experiment on English--French, English--German, English--Spanish and English--Italian. We filter the parallel corpus for each language pair by removing sentences that cover tokens out of 2 languages. Raw and filtered number of the parallel sentences for each pair are shown in Table~\ref{data}. 10,000 sentences are selected for validation on each language pair. We tokenize sentences by SentencePiece\footnote{\url{https://github.com/google/sentencepiece}}~\cite{kudo-2018-subword} and build a shared vocabulary with the size of 50k for each language pair. 

For each encoder, we use the transformer architecture with 2 hidden layers, 8 attention heads, hidden size of 512 and filter size of 1,024, and the parameters of two encoders are shared with each other. The sentence representations generated are 512 dimensional. For the training phase, it minimizes the weighted losses for our proposed cross-lingual language model jointly with 2 auxiliary tasks. We train 12 epochs for each language pair (30 epochs for English-Italian because of nearly half number of parallel sentences) with the Adam optimizer, learning rate of 0.001 with warm-up strategy for 3 epochs (6 epochs for English-Italian) and dropout-probability of 0.1 on a single TITAN X Pascal GPU with the batch size of 128 paired sentences. Training loss for each language pair can converge within 10 GPU (12GB)$\times$days, which is far more efficient than most cross-lingual sentence representation learning methods.\footnote{Note that it is impractical to compare the efficiency with LASER, which is trained by 80 V100 GPU$\times$days due to different training data settings. However, it is obvious that our lightweight model is significantly more efficient than the 5-layer LSTM-based encoder-decoder model structure of LASER, because of the parallel computing nature of the transformer encoder \cite{DBLP:conf/nips/VaswaniSPUJGKP17} of our model without any decoder.}

\subsection{Baselines}
\label{sec:baeslines}
For evaluation on the MLDoc benchmark, we use the state-of-the-art fixed-dimensional word representation methods MultiCCA+CNN method~\cite{schwenk-li-2018-corpus} and Bi-Sent2Vec~\cite{DBLP:journals/corr/abs-1910-01108}, the representative fixed-dimensional sentence representation methods~\cite{yu-etal-2018-multilingual}, LASER \cite{artetxe-schwenk-2019-massively}, and T-LASER \cite{DBLP:journals/corr/abs-2008-08567} as baselines. In addition, as reference only, we present the results of the global fine-tuning methods, mBERT~\cite{devlin-etal-2019-bert} and the state-of-the-art BERT-based variant, MultiFit~\cite{eisenschlos-etal-2019-multifit}. 

For the XSR task, bilingual fixed-dimensional methods, Bi-Vec~\cite{luong-etal-2015-bilingual} \& Bi-Sent2Vec ~\cite{DBLP:journals/corr/abs-1912-12481}, and multilingual fixed-dimensional methods, TransGram~\cite{coulmance-etal-2015-trans} \& LASER~\cite{artetxe-schwenk-2019-massively} are used as baselines. 

Note that T-LASER and LASER are trained on 223M parallel sentences on 93 languages, which uses significantly more training data than ours.

We also show the results by comparing with~\cite{reimers-gurevych-2020-making} in Appendix~\ref{app}, which is a recent work using global fine-tuning methods to generate multilingual sentence representations.

\subsection{MLDoc: Zero-shot Cross-lingual Document Classification}
The MLDoc task, which consists of news documents given in 8 different languages, is a benchmark to evaluate cross-lingual sentence representations. We conduct our evaluations in a zero-shot scenario: we train and validate a new linear classifier on the top of the pre-trained sentence representations in the source language, and then evaluate the classifier on the test set for the target language. We implement the evaluation by facebook's MLDoc library.\footnote{\url{https://github.com/facebookresearch/MLDoc}} As shown in Table~\ref{mldoc}, our lightweight transformer model obtains the best results for most language pairs compared with previous fixed-dimensional word and sentence representation learning methods. 
Our methods yield only slightly worse performance even when compared with the state-of-the-art global fine-tuning style method, MultiFit~\cite{eisenschlos-etal-2019-multifit}, on this task. This is because the entire model will be updated in the fine-tuning phase, which indicates more parameters will be task-specific after fine-tuning. For fixed-dimensional methods, just an additional dense layer will be trained, which leads to their higher efficiency. 


\begin{table*}[t]
\begin{center}
\begin{tabular}{lrr|rrrr|rrrr}
      \toprule
      \multirow{2}{*}{N} & \multirow{2}{*}{M}& \multirow{2}{*}{T} &\multicolumn{4}{c|}{MLDoc} & \multicolumn{4}{c}{XSR} \\
      &&&en$\rightarrow$fr&fr$\rightarrow$en&en$\rightarrow$es&es$\rightarrow$en&en$\rightarrow$fr&fr$\rightarrow$en&en$\rightarrow$es&es$\rightarrow$en\\
      \midrule
      1 & 7,135 & 19 &81.7&79.4 & 75.5 & 74.9 & 89.4&90.0 & 86.4 & 87.7 \\
      \textbf{2} & \textbf{11,607} & \textbf{24} &\textbf{85.1}&\textbf{82.4}& 80.8 & \textbf{79.2} &90.2&90.8 & 90.7 & 91.2\\
      3 & 16,804 & 29 &84.2&81.9 & \textbf{81.2} & 78.1 & 90.9&91.5 & 91.1 & 92.0 \\
      4 & 21,923 & 34 &84.2&82.0 & 81.1 & 78.7 & \textbf{91.4}&91.5 & 91.5 & \textbf{92.2} \\
      6 & 28,024 & 44 &83.0&80.8 & 79.8 & 78.3 & 91.2&\textbf{92.0} & \textbf{91.7} & 91.9\\
      \bottomrule
\end{tabular}
\caption{\textbf{Training efficiencies with different numbers of layers.} N denotes number of layers within the transformer encoder; M and T indicate memory overhead (MB) and training time (min), respectively. Memory overhead changes for different languages and here we report the numbers on English--French. Training time is measured every 10,000 training steps. The results are reported by using a single V100 GPU card with the batch size of 128 sentences. 2-layer is the default setting for our lightweight model.}
\label{eff1}

 \end{center}
\end{table*}

\begin{table*}[t]
\begin{center}
\begin{tabular}{l|rrrr|rrrr}
      \toprule
      \multirow{2}{*}{Tasks} & \multicolumn{4}{c|}{MLDoc} & \multicolumn{4}{c}{XSR} \\
      &en$\rightarrow$fr&fr$\rightarrow$en&en$\rightarrow$es&es$\rightarrow$en&en$\rightarrow$fr&fr$\rightarrow$en&en$\rightarrow$es&es$\rightarrow$en\\
      \midrule
      MLM & 78.5 & 77.6 & 74.6 & 75.9 & 19.6 & 25.4 & 11.2 & 28.5 \\
      SMLM & 75.0 & 78.7 & 75.3 & 74.0 & 85.0 & 85.3 & 86.4 & 87.1 \\
      XTR & 84.2 & 81.2 & 79.9 & 77.6 & 89.5 & \textbf{90.8} & \textbf{90.3} & 89.5\\
      MLM $\oplus$ XTR & 82.2 & 78.2 & 78.4 & 76.7 & 84.1 & 85.0 & 87.6 & 88.9 \\
      UGT (SMLM $\oplus$ XTR) & \textbf{85.1} & \textbf{82.4} & \textbf{80.8} & \textbf{79.2} & \textbf{89.8} & 90.6 & 89.4 & \textbf{89.6} \\
      \bottomrule
\end{tabular}
\caption{\textbf{Effectiveness of different generative tasks.} UGT indicates ``SMLM $\oplus$ XTR", which indicates the training task combining SMLM and XTR. MLM $\oplus$ XTR denotes the unified training task combining MLM and XTR.} 
\label{lm}

 \end{center}
\end{table*}

\begin{table*}[t]
\begin{center}
\begin{tabular}{l|rrrr|rrrr}
      \toprule
      \multirow{2}{*}{Tasks} & \multicolumn{4}{c|}{MLDoc} & \multicolumn{4}{c}{XSR} \\
      &en$\rightarrow$fr&fr$\rightarrow$en&en$\rightarrow$es&es$\rightarrow$en&en$\rightarrow$fr&fr$\rightarrow$en&en$\rightarrow$es&es$\rightarrow$en\\
      \midrule
      UGT & \textbf{85.1} & \textbf{82.4} & \textbf{80.8} & \textbf{79.2} & 89.8 & 90.6 & 89.4 & 89.6 \\
      + align & 84.1 & 81.9 & 78.7 & 77.9 & 89.9 & 90.4 & 89.8 & 90.9 \\
      + align + sim & 82.3 & 80.3 & 77.6 & 76.2 & \textbf{90.2} & \textbf{90.8} & \textbf{90.7} & \textbf{91.2} \\
      \bottomrule
\end{tabular}
\caption{\textbf{Effectiveness of the contrastive tasks.} UGT indicates the training without any sentence-level contrastive tasks.} 
\label{csloss}

 \end{center}
\end{table*}

\subsection{XSR: Cross-lingual Sentence Retrieval}
We also conduct an evaluation to gauge the quality of our cross-lingual sentence representations on the bitext mining task, which is identical to some components of the training task. Specifically, given 2,000 sentences in the source language, we conduct the corresponding sentence retrieval from 200K sentences in the target language. P@1 scores of our lightweight models
and previous bilingual representation methods calculated by~\citet{artetxe-schwenk-2019-margin} are reported. As shown in Table~\ref{csr}, we observe that our lightweight models outperform the bilingual pooling-based representation learning methods by a significant margin, which reflects the basic ability of the contextualized representations generated by our lightweight models. 
However, our lightweight models underperform 
LASER, which can be attributed to our lightweight capacities and bilingual settings. Note that LASER uses significantly larger multilingual training data (see Section~\ref{sec:baeslines}).

\subsection{Analyses}
\label{LMS}

We perform ablation experiments to confirm the efficiency and the effectiveness of each training task for our models. Analyses for other hyperparameter configurations of batch size, sentence representation dimension, and training corpus size are presented in Appendix~\ref{app}.

\noindent{\textbf{Relation among Number of Layers, Efficiency, and Performances.}}
We report the efficiency statistics and performances of our proposed methods trained by different layer number settings. As shown in Table~\ref{eff1}, we observe a linear increase of memory occupation and training time per 10,000 training steps by increasing the number of transformer encoder layers. Specifically, a 6-layer transformer encoder occupies nearly 2.5 times memory and costs 1.8 times training time compared to our 2-layer model. Therefore, given the same memory occupation (by adjusting the batch size), theoretically our lightweight model can be implemented over 4 times ($\approx2.5\times1.8$) faster than the 6-layer model. Concerning the respective performances on MLDoc and XSR, we see that lightweight model with 2 transformer layers obtains the peak performance on MLdoc, and the performances decrease when we add more layers. This indicates that the 2-layer transformer encoder is an ideal structure for our proposed training tasks on the document classification task. On the other hand, performances on XSR keep increasing gradually with more layers, where the 1-layer model can even yield decent performance on this task.

Our proposed training tasks perform well from the 2-layer model, while 6 layers are required for standard MLM and 5 LSTM layers are required for LASER. This is why we use 2-layer as the basic unit for our model.

\noindent{\textbf{Effectiveness of Different Generative Tasks.}} We report the results with different generative tasks in Table~\ref{lm}. We observe that XTR outperforms other generative tasks by a significant margin on both MLDoc and XSR downstream tasks. XTR yields further improvements when unified with SMLM, which is introduced as the generative task in our model. This demonstrates the necessity of a well-designed generative task for the lightweight dual-transformer architecture. 

\noindent{\textbf{Effectiveness of the Contrastive Tasks.}} In Table~\ref{csloss}, we study the contribution of the sentence-level contrastive tasks. We observe that a higher performance on MLDoc is yielded by the vanilla model while more sentence-level contrastive tasks improve the performance on XSR. This can be attributed to the similar nature between the supervision provided by sentence-level contrastive tasks and XSR process. In other words, contrastive-style tasks have a detrimental effect on the document classification downstream task. In future work, we will explore how to train a balanced sentence representation model with contrastive tasks.

\section{Conclusion}
In this paper, we presented a lightweight dual-transformer based cross-lingual sentence representation learning method. For the fixed 2-layer dual-transformer framework, we explored several generative and contrastive tasks to ensure the sentence representation quality and facilitate the improvement of the training efficiency. In spite of the lightweight model capacity, we reported substantial improvements on MLDoc compared to fixed-dimensional representation methods and we obtained comparable results on XSR. In the future, we plan to verify whether our proposed methods can be combined with knowledge distillation.


\section*{Acknowledgements}
We would like to thank all the reviewers for their valuable comments and suggestions to improve this paper. This work was partially supported by Grant-in-Aid for Young Scientists \#19K20343, JSPS.

\bibliographystyle{acl_natbib}
\bibliography{anthology,acl2021}

\clearpage
\appendix
\section{Appendices}
\label{app}

\noindent{\textbf{Comparisons with} \citet{reimers-gurevych-2020-making}\textbf{.}} \citet{reimers-gurevych-2020-making} use the knowledge distillation to train multilingual sentence embeddings where pre-trained encoders are utilized to initialize the teacher and student model, which is a kind of the \textit{global fine-tuning style methods} fine-tuned by parallel sentences. As the results on MLDoc and XSR shown in Table~\ref{making}, their multilingual representations yield good performance on bitext mining but perform poorly on classification tasks. This demonstrates the importance of exploring task-agnostic multilingual sentence representations like LASER and ours.

\begin{table}[t]
\begin{center}
\resizebox{\linewidth}{!}{  
\begin{tabular}{lrrrr}
      \toprule
      \multirow{2}{*}{Batch Size} & \multicolumn{2}{c}{MLDoc} & \multicolumn{2}{c}{XSR} \\
      &en$\rightarrow$fr&fr$\rightarrow$en&en$\rightarrow$fr&fr$\rightarrow$en\\
      \midrule
      64 & 82.9 & \textbf{82.6} & 89.6 & 90.3 \\
      128 & \textbf{84.1} & 81.9 & \textbf{90.2} & \textbf{90.8} \\
      256 & 82.9 & 81.1 & \textbf{90.2} & 90.7 \\
      \bottomrule
\end{tabular}
}
\caption{Effect of the batch size.} 
\label{bs}

 \end{center}
\end{table}

\begin{table}[t]
\begin{center}
\resizebox{\linewidth}{!}{  
\begin{tabular}{lrrrr}
      \toprule
      \multirow{2}{*}{Corpus Size} & \multicolumn{2}{c}{MLDoc} & \multicolumn{2}{c}{XSR} \\
      &en$\rightarrow$fr&fr$\rightarrow$en&en$\rightarrow$fr&fr$\rightarrow$en\\
      \midrule
      12.5\% &82.5&80.7&\textbf{90.8}&90.5\\
      25\% &82.5&80.3&90.5&\textbf{91.2}\\
      50\% &83.0&81.5&90.2&91.0\\
      100\% & \textbf{85.1} & \textbf{82.4} & 90.2 & 90.8 \\
      \bottomrule
\end{tabular}
}
\caption{Impact of the corpus size.} 
\label{cs}

 \end{center}
\end{table}

\noindent{\textbf{Batch Size.}} We investigate the effect of the batch size for contrastive tasks, where different batch sizes indicate the discrepancy of the negative sample numbers. As shown in Table~\ref{bs}, larger batch harms the lightweight model based sentence representation learning and 128 is reported as the best batch size setting for our lightweight model. Furthermore, batch size of 128 allows the training to be assigned on 12GB GPU card while a larger batch size requires more GPU memory.

\begin{figure}[t]
\begin{center}
\includegraphics[width=\linewidth]{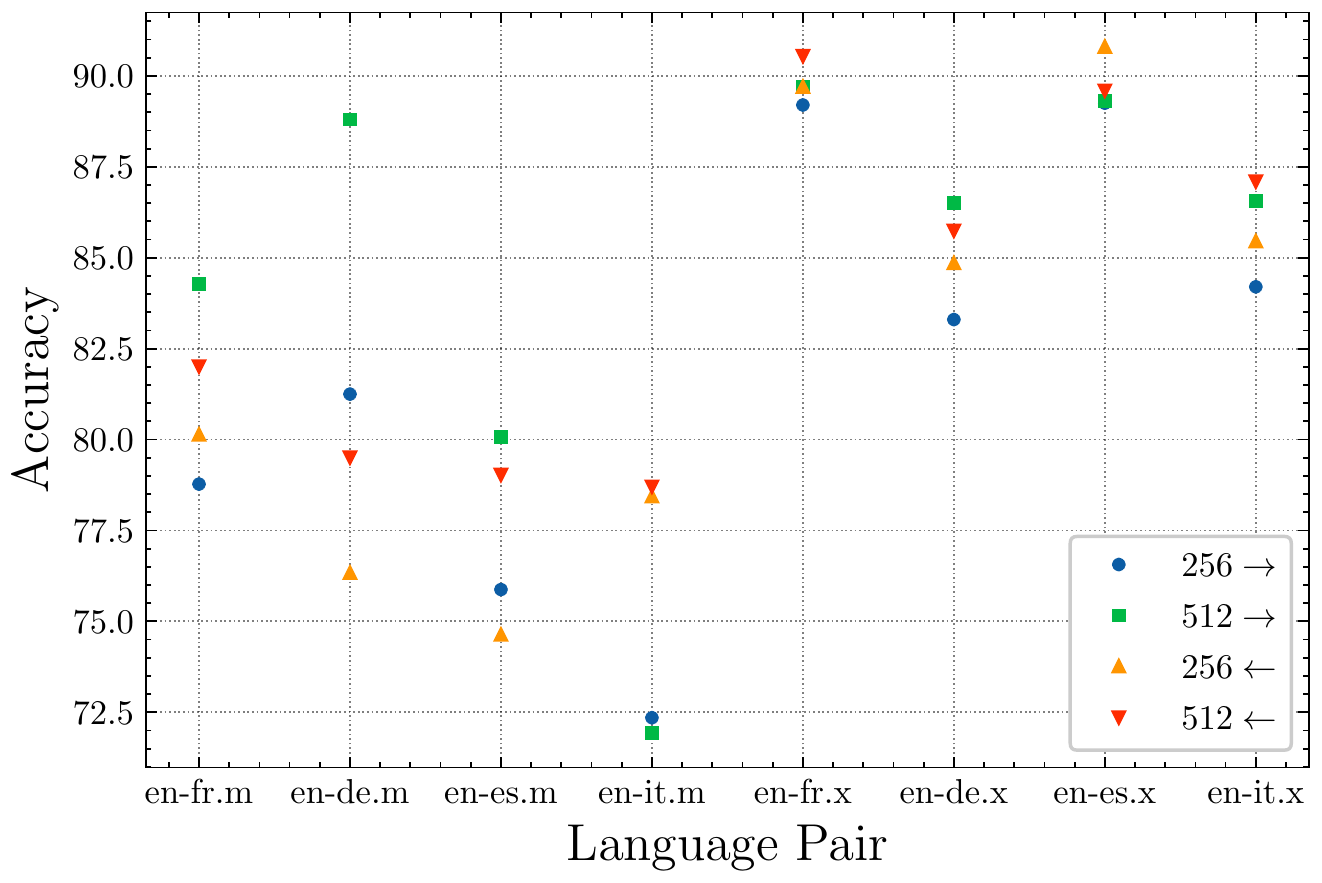}
\caption{\textbf{Performance of different representation dimensions on MLDoc (.m) and XSR (.x).} Arrows denotes direction of zero-shot setting.}
\label{emb_mldoc}
\end{center}
\end{figure}

\noindent{\textbf{Corpus Size.}} We show the impact of the size of the parallel corpus on English-French in Table~\ref{cs}. For MLDoc, we observe higher accuracy on larger corpus while for XSR, a small fraction of the large corpus suffices to yield effective results. This indicates that more parallel data improves the performance on MLDoc.

\noindent{\textbf{Sentence Representation Dimension.}} In Figure~\ref{emb_mldoc}, we present the effect of the sentence representation dimension. 512-dimensional sentence representations significantly outperform 256-dimensional ones in our lightweight model. Moreover, representation size of 512 yields better performance without increasing the training time.

\begin{table*}[t!]
\begin{center}
\resizebox{\linewidth}{!}{  
\begin{tabular}{lr rrrr rrr r}
      \toprule
      \multirow{2}{*}{Method}&\multicolumn{2}{c}{en-fr}&\multicolumn{2}{c}{en-de}&\multicolumn{2}{c}{en-es}&\multicolumn{2}{c}{en-it}&\multirow{2}{*}{Avg.}\\
      &$\rightarrow$&$\leftarrow$&$\rightarrow$&$\leftarrow$&$\rightarrow$&$\leftarrow$&$\rightarrow$&$\leftarrow$&\\
      \toprule
      \textit{MLDoc} \\
      \citet{reimers-gurevych-2020-making} & 68.0 & 78.5 & 77.6 & 79.2 & 72.7 & 72.2 & 68.5 & 74.2 & 73.9 \\
      Ours &85.1&82.4&88.8&80.8&80.8&79.2&74.3&79.9&81.4\\
      \hline
      \hline
      \textit{XSR} \\
      \citet{reimers-gurevych-2020-making} & 93.0 & 92.3 & 89.9 & 89.2 & 93.9 & 92.9 & 91.7 & 91.4 & 91.8 \\
      Ours & 90.2 & 90.8	&86.3	&86.9&90.7&91.2&86.9&87.6&88.8\\
      \bottomrule
\end{tabular}
}
\caption{Comparisons with \citet{reimers-gurevych-2020-making} on MLDoc and XSR.}
\label{making}

 \end{center}
\end{table*}


\end{document}